\documentclass[twoside]{article}

\usepackage[accepted]{aistats2026}
\usepackage{amsmath, amssymb, mathtools, bbm, graphicx, booktabs, multirow, url}
\usepackage[dvipsnames]{xcolor}

\usepackage[round]{natbib}

\usepackage{hyperref}

\newcommand{\E}{\mathbb{E}}
\renewcommand{\P}{\mathbb{P}}
\newcommand{\R}{\mathbb{R}}

\newcommand{\cPY}{\Delta_k}

\newcommand{\myS}{\operatorname{S}}

\newcommand{\BS}{\operatorname{BS}}
\newcommand{\MCB}{{\operatorname{MCB}}}
\newcommand{\DSC}{{\operatorname{DSC}}}
\newcommand{\UNC}{{\operatorname{UNC}}}
\newcommand{\CE}{{\operatorname{CE}}}

\newcommand{\one}{\mathbbm{1}}

\begin{document}

%

%

\twocolumn[

\aistatstitle{Calibration in Context: A Case Study with Score Decompositions}

\aistatsauthor{Johannes Resin}

\aistatsaddress{Goethe University Frankfurt, Germany} ]

\begin{abstract}
	Decompositions of proper scores into measures of miscalibration (reliability), discrimination (resolution), and uncertainty have a long history in weather forecasting. In machine learning (ML), related calibration error metrics are now seeing a surge of interest. In this note, I review the close connection between these concepts and present a small case study on image classifiers from the literature.
	The study exemplifies that an exclusive focus on calibration error may lead to questionable conclusions when improvements in calibration come at the expense of a drastic decline in overall predictive performance. I critically examine histogram binning and show that isotonic regression produces better overall recalibration results. A simple linear interpolation of the isotonic fit is shown to further improve predictive performance without loss of calibration.
\end{abstract}

\section{BACKGROUND \& MOTIVATION}

By now, the need for reliable uncertainty quantification is widely recognized in the ML literature \citep[e.g.,][]{HW21}. Probabilistic classifiers address this need by predicting class distributions. Calibration---the statistical compatibility between predictive distributions and true classes---is crucial for decision-making and trustworthy AI systems \citep[e.g.,][]{VWA+19}.
The predictive performance of such classifiers is typically assessed using proper scoring rules \citep{GR07}, which can be decomposed into miscalibration (reliability), discrimination (resolution), and uncertainty components
\citep[e.g.,][]{Bro09, Pohle2020}, building on the seminal work of \citet{Sanders1963} and \citet{Murphy73}. In contrast, many authors in ML advocate the use of calibration error metrics \citep[e.g.,][]{NCH15, GPW17, KLM19, NDZ+19, GR22, BGHN23, RSB+25}, often treating them as primary evaluation criteria.  

This note highlights the importance of considering calibration in the context of overall predictive performance, critically examining a finding by \citet{GR22}.
The study aligns with recent research that appreciates score decompositions and addresses the interplay and tensions between calibration and overall predictive ability \citep{GB22,Silva2023,MCF+24,PGS+24,BHJB25,CG25}.
A recent strand of statistical literature suggests the use of non-parametric isotonic regression as a robust alternative to traditional binning-based estimators of score components. Here, I adapt this approach to (probabilistic) multi-class classifiers, leveraging work on binary classifiers \citep{DGJ21}, and point and distributional forecasts for real-valued outcomes \citep{EJS23,AWGZ24}.
The decomposition is used to study class-wise recalibration approaches based on histogram binning and isotonic regression. The study shows that a simple smoothing of isotonic fits preserves discrimination effectively resulting in the best overall performance.

\section{SETTING \& METHODS}

This section briefly introduces the problem setup and key tools for quantitative out-of-sample verification.

\subsection{Basic Setting}

Classification aims to predict the class label $Y \in \{1,\dots,k\}$ of an instance characterized by a feature vector $X \in \R^d$. A probabilistic classifier $c\colon \R^d\rightarrow\cPY = \{p = (p_j)_{j = 1}^k \in [0,1]^k \mid \sum_j p_j = 1\}$ maps the features $X$ to a probability distribution given by a vector of class probabilities from the probability simplex $\cPY$.
The variables $X$ and $Y$ are assumed to follow a joint distribution $\P$ on a probability space $\Omega$, where $\P_{Y\mid c(X)}$ denotes the conditional distribution of $Y$ given the prediction $c(X)$, and $\P_{Y}$ denotes the marginal distribution of $Y$. Ideally, the classifier $c$ should be calibrated \citep{VWA+19} in the sense that
\begin{equation}\label{Eq:AC}
	\P_{Y\mid c(X)} = c(X),
\end{equation}
while also effectively discriminating between the different classes. Together, calibration and discrimination are the key properties that drive the overall predictive performance of a classifier.

\subsection{Recalibration Methods}
\label{Sec:Recal}

To improve calibration, classifiers are often recalibrated using various methods \citep[see][]{Silva2023}. The common class-wise approach recalibrates each class probability individually using a binary calibrator and then normalizes the resulting probabilities \citep{ZE02}. However, \citet{GR22} have recently suggested to omit the normalization step.
The proposal raises significant conceptual and practical concerns: \emph{How can binary class probabilities that do not sum to one be meaningfully interpreted or used for effective decision-making?}\/ Rather than pursue these issues further, I reexamine the findings by \citet{GR22} empirically in this note.

\begin{figure}[b]
	\centering
	\includegraphics[width=0.8\columnwidth]{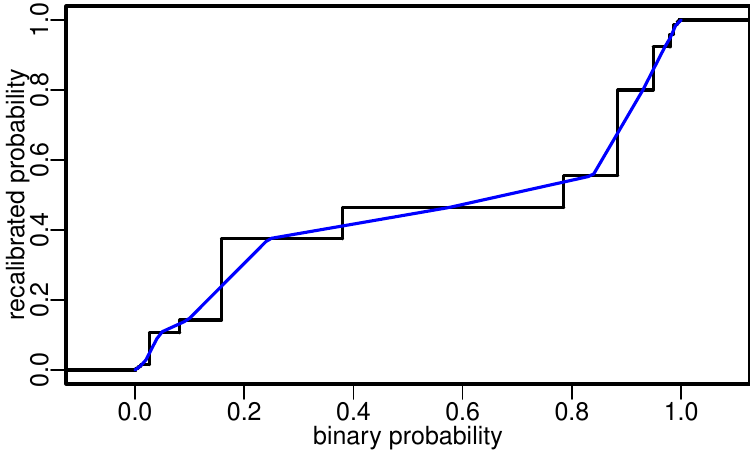}
	\caption{Isotonic fit (black) and smoothing via linear interpolation anchored at bin medians (\textcolor{blue}{blue}).}\label{Fig:SmoothIR}
\end{figure}

The subsequent case study focuses on the class-wise approach, contrasting the use of histogram binning as used by \citet{GR22} and isotonic regression. Histogram binning partitions the unit interval into a fixed number of bins, recalibrating predictions in each bin to match the conditional event frequency of the bin. This method requires some tuning of the type of binning (equal-width or equal mass) and number of bins. In contrast, isotonic regression \citep{Barlow1972} is a non-parametric, tuning-free method that finds (in-sample) optimal predictions under a natural monotonicity constraint via the pool-adjacent-violators algorithm. The isotonic fit is characterized by its piecewise constant segments, which give rise to a data-driven binning.
The case study also includes a simple smoothing of the isotonic fit that linearly interpolates between bin medians, as described in Appendix \ref{App:SIR} and illustrated in Figure \ref{Fig:SmoothIR}, to avoid discretizing recalibrated probabilities and preserve discrimination more effectively. This approach is closely related to the method by \citet{JOKO11}, who propose a piecewise cubic Hermite spline interpolation that produces a more refined smoothing.

\subsection{Proper Scoring Rules}

Scoring rules quantify overall predictive performance by assigning numeric scores to prediction-observation pairs $(p,y) \in \cPY \times \{1,\dots,k\}$. A scoring rule $\myS$ is \emph{strictly proper}\/ if the expected score is always minimized by the true class distribution, i.e.,
\[
	\E_{Y\sim p} \myS(p,Y) < \E_{Y\sim p} \myS(q,Y) \text{ for all } p, q \in \cPY, p\neq q.
\]
Here, I use the \emph{Brier score}---a popular strictly proper scoring rule dating back to \citet{Brier1950}---given by 
\begin{equation}\label{Eq:BS}
	\BS(p,y) = \sum_{j = 1}^k (\one\{y = j\} - p_j)^2
\end{equation}
with indicator $\one\{y = j\}$ equal to 1 if $y = j$ and 0 otherwise. The Brier score is a sum of binary Brier scores $(p_j, \one\{y = j\}) \mapsto (\one\{y = j\} - p_j)^2$ for each class $j$.

\subsection{Score Decompositions}

The expected score $\overline{\myS} = \E \myS(c(X),Y)$ under a proper scoring rule can be decomposed as 
\begin{equation}\label{Eq:Decomp}
	\overline{\myS} = \MCB - \DSC + \UNC,
\end{equation}
where $\MCB = \overline{\myS} - \E \myS(\P_{Y\mid c(X)},Y)$ quantifies \emph{miscalibration}, $\DSC = \E \myS(\P_{Y},Y) - \E \myS(\P_{Y\mid c(X)},Y)$ quantifies \emph{discrimination}, and $\UNC = \E \myS(\P_{Y},Y)$ (uncertainty) is a reference score attained by the marginal distribution (which is calibrated but does not discriminate at all).

\subsection{Calibration Error}

Calibration error is typically defined as the expected distance between predicted and true conditional distribution under a distance $d \colon \cPY^2 \rightarrow [0,\infty)$, i.e., 
\[
	\CE = \E d(c(X),\P_{Y\mid c(X)}).
\]
When $d(q,p) = \E_{Y \sim p} [\myS(q,Y) - \myS(p,Y)]$ is the Bregman divergence associated with a proper scoring rule $\myS$ (see \citealp{GR07}), a simple calculation
confirms that $\CE = \MCB$. In what follows, I focus on the Brier score decomposition with its miscalibration component as calibration error.

\subsection{Estimation of Score Components}

In practice, the decomposition \eqref{Eq:Decomp} is estimated on a test set of $n$ prediction-observation pairs $(p_1,y_1), \dots, (p_n,y_n)$, requiring estimates $\widehat{p}_1,\dots,\widehat{p}_n$ of the conditional distribution $\P_{Y\mid c(X) = p_i}$ given prediction $i = 1,\dots,n$. Using the empirical class frequencies $\widehat{p}_0$ to estimate the marginal distribution, an empirical decomposition of the average score $\widehat{\myS} = \frac1n\sum_{i = 1}^n \myS(p_i,y_i)$ is obtained as
\begin{align}
\widehat{\myS} &= \widehat{\MCB} - \widehat{\DSC} + \widehat{\UNC}, \label{Eq:EmpDecomp} \\
\text{where }~\widehat{\MCB} &= \widehat{\myS} - \textstyle\frac1n\sum_i \myS(\widehat{p}_i,y_i), \notag \\
\widehat{\DSC} &= \textstyle\frac1n\sum_i \myS(\widehat{p}_0,y_i) - \myS(\widehat{p}_i,y_i), \text{ and} \notag \\ 
\widehat{\UNC} &= \textstyle\frac1n\sum_i \myS(\widehat{p}_0,y_i). \notag
\end{align}
Similar to the Brier score based decomposition of the CRPS in \citet{AWGZ24}, the structure of the Brier score in \eqref{Eq:BS} admits directly applying the decomposition by \citet{DGJ21} to the binary summands. This approach yields a \emph{class-wise}\/ decomposition and is equivalent to fitting an isotonic regression separately for each class, without normalizing the recalibrated class probabilities. Unlike standard binning estimators, the isotonic approach is independent of implementation choices while being based on a consistent estimator \citep{DGJ21}. However, the class-wise decomposition quantifies deviations from a weaker form of calibration, namely, \emph{class-wise calibration}\/ \citep{GR22} given by 
\[
	\P(Y = j \mid p_j) = p_j \quad\text{for all}\quad j \in \{1,\dots,k\}.
\]
Appendix \ref{App:SDwN} contains additional results for an alternative \emph{normalized}\/ decomposition that uses isotonic regression with a normalization step to estimate the conditional class probabilities $\widehat{p}_i$ to be plugged into \eqref{Eq:EmpDecomp}. The normalized decomposition quantifies deviations from the stronger form of calibration in \eqref{Eq:AC}.

\section{CASE STUDY}
\label{Sec:Study}

\begin{table*}[ht]
	\caption{Class-wise Brier score decomposition ($\times 100$) and classification accuracy (Acc) averaged across seven base models for both datasets (CIFAR-10 and CIFAR-100) and each recalibration method. The best average values per metric are highlighted in \textbf{bold}. Values in \textcolor{BrickRed}{red} indicate a drastic decline in predictive performance.}\label{Tab:avgSD}
	\centering
	\begin{tabular}{lccccc}
		\toprule
		\multicolumn{6}{c}{CIFAR-10} \\
		\midrule
		Method & $\BS$ & $\MCB$ & $\DSC$ & $\UNC$ & Acc \\ 
		\midrule
		base & 0.781 & 0.063 & 8.282 & 9 & \textbf{0.951} \\ 
		CW-HB & \textcolor{BrickRed}{1.832} & \textbf{0.009} & \textcolor{BrickRed}{7.177} & 9 & \textcolor{BrickRed}{0.890} \\ 
		N-HB & \textcolor{BrickRed}{1.237} & 0.070 & \textcolor{BrickRed}{7.833} & 9 & \textcolor{BrickRed}{0.890} \\ 
		CW-IR & 0.779 & 0.027 & 8.248 & 9 & 0.950 \\ 
		N-IR & 0.770 & 0.047 & 8.277 & 9 & 0.950 \\ 
		CW-SIR & 0.769 & 0.051 & 8.282 & 9 & 0.950 \\ 
		N-SIR & \textbf{0.764} & 0.047 & \textbf{8.283} & 9 & 0.950 \\ 
		\bottomrule
	\end{tabular}~~
	\begin{tabular}{lccccc}
		\toprule
		\multicolumn{6}{c}{CIFAR-100} \\
		\midrule
		Method & $\BS$ & $\MCB$ & $\DSC$ & $\UNC$ & Acc \\ 
		\midrule
		base & 0.327 & 0.039 & 0.702 & 0.99 & \textbf{0.775} \\ 
		CW-HB & \textcolor{BrickRed}{0.865} & \textbf{0.003} & \textcolor{BrickRed}{0.128} & 0.99 & \textcolor{BrickRed}{0.483} \\ 
		N-HB & \textcolor{BrickRed}{0.532} & 0.036 & \textcolor{BrickRed}{0.494} & 0.99 & \textcolor{BrickRed}{0.483} \\ 
		CW-IR & 0.333 & 0.021 & 0.677 & 0.99 & 0.770 \\ 
		N-IR & 0.324 & 0.034 & 0.699 & 0.99 & 0.770 \\ 
		CW-SIR & 0.323 & 0.035 & 0.702 & 0.99 & 0.773 \\ 
		N-SIR & \textbf{0.320} & 0.033 & \textbf{0.703} & 0.99 & 0.773 \\ 
		\bottomrule
	\end{tabular}
\end{table*}

\begin{figure*}[t]
	\centering
	\includegraphics[width = 0.49\textwidth]{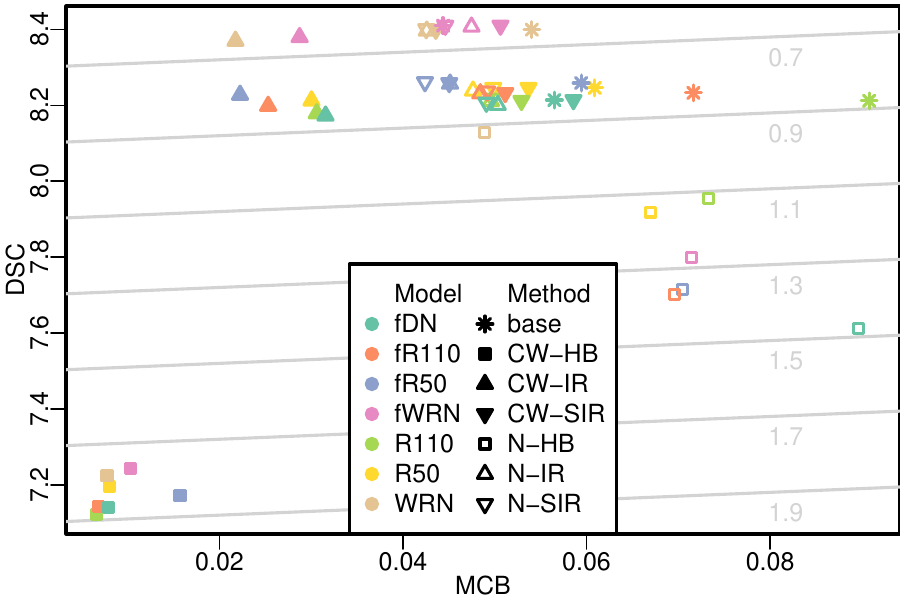}
	\includegraphics[width = 0.49\textwidth]{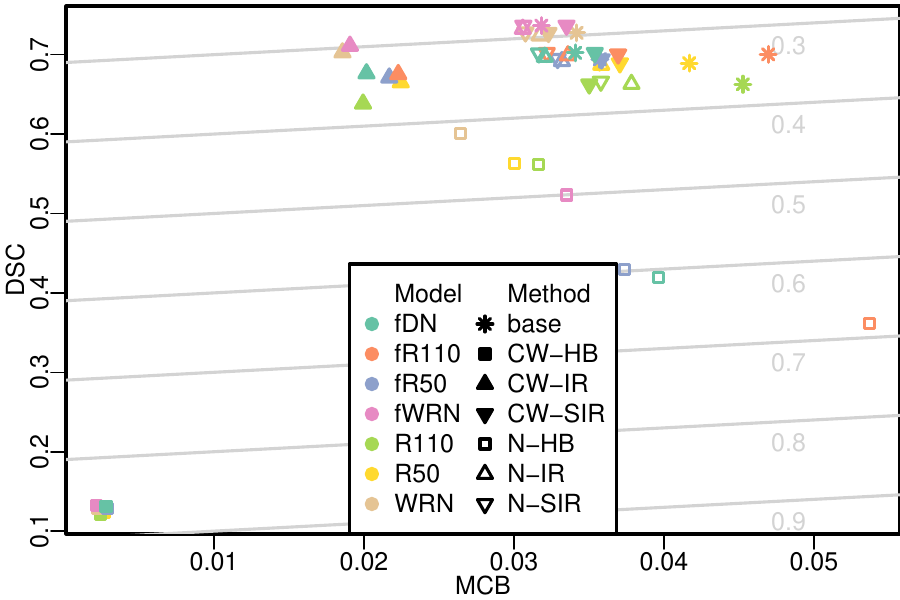}
	\caption{MCB-DSC plots for CIFAR-10 (left) and CIFAR-100 (right) showing class-wise Brier score decompositions ($\times 100$). The plots show results for seven base models that differ in architecture (\textbf{R}esNet-\textbf{50}, \textbf{R}esNet-\textbf{110}, \textbf{W}ide\textbf{R}es\textbf{N}et, and \textbf{D}ense\textbf{N}et-121) and training loss (\textbf{f}ocal loss or Brier score). The \textcolor{gray}{gray} isolines correspond to MCB-DSC pairs that yield the same overall Brier score.}
	\label{Fig:MCB-DSC}
\end{figure*}

The case study builds upon an experiment from \citet[][]{GR22}, who compared several recalibration methods for common image classifiers using a custom calibration error metric (CW-ECE) on CIFAR-10 and CIFAR-100 \citep{CIFAR}. In their study, they do not use other metrics such as proper scoring rules or classification accuracy. They propose omitting the standard normalization step in class-wise recalibration with binary histogram binning (CW-HB), contrasting it with the usual normalized approach (N-HB). \citet[p.\ 9, bold in original]{GR22} report the following finding: 
\begin{quote}
	``For CW-ECE, CW-HB is the best performing method across the two datasets and all four architectures. The N-HB method which has been used in many CW-ECE baseline experiments performs terribly. In other words, skipping the normalization step leads to a large improvement	in CW-ECE. \textbf{This observation is one of our most striking findings.}''
\end{quote}
Here, I reexamine this finding via the class-wise Brier score decomposition.
The case study uses calibration and test sets of raw model predictions from the supplementary material for \citet{GR22}. The datasets include predictions from seven base models \citep{Zagoruyko2016, Huang2017, Mukhoti2020}: ResNet-50, ResNet-110, WideResNet (each trained with either Brier score or focal loss), and DenseNet-121 (trained with focal loss). The present study features the following recalibration methods: histogram binning with normalization (N-HB) and without normalization (CW-HB) as used by \citet{GR22}, isotonic regression with and without normalization (N-IR and CW-IR, respectively), and the smoothed version of isotonic regression (N-SIR and CW-SIR) introduced in Section \ref{Sec:Recal}.

Table \ref{Tab:avgSD} presents the score decompositions along with classification accuracy, i.e., the frequency with which the predicted (most likely) class matches the true class, averaged across all base models for each recalibration method and the two datasets.
The MCB-DSC plots in Figure \ref{Fig:MCB-DSC} provide a visualization of the individual performance for each model and recalibration method by plotting discrimination, $\DSC$, against miscalibration, $\MCB$. The isolines on these plots represent decompositions with equal overall predictive performance, illustrating trade-offs between miscalibration and discrimination. Similar plots have appeared in \citet{ARSIA23, AWGZ24, DGJV24}. Analogous tables and plots showing a normalized version of the Brier score decomposition are presented in Appendix \ref{App:SDwN}.

In line with the CW-ECE results reported by \citet{GR22}, the class-wise decomposition yields a low calibration error, $\MCB$, for CW-HB. However, this improvement comes at the cost of a drastic decline in discrimination, $\DSC$, resulting in a doubled Brier score and twice as many misclassifications (1 - Acc). While the normalization step in N-HB partially recovers discrimination ability, predictive performance remains substantially reduced. In contrast, the normalized score decomposition (presented in the appendix) attributes most of the score difference between CW-HB and N-HB to a large miscalibration component for CW-HB, because this decomposition quantifies deviations from the stronger notion of calibration in \eqref{Eq:AC}. Apart from this exception, the normalized decomposition shows fairly similar results with a slight increase in $\MCB$ for the class-wise (CW-X) approaches and a slight decrease in $\MCB$ for the normalized (N-X) versions. While CW-IR exhibits lower miscalibration than the other methods based on isotonic regression, this improvement comes at the cost of a moderate loss of discrimination, resulting in overall performance similar to the base classifier. Notably, the smoothed version CW-SIR does not show this trade-off, which may thus be due to CW-IR discretizing the recalibrated class probabilities. Ultimately, smoothed isotonic regression with normalization (N-SIR) achieves the best overall performance, reducing miscalibration while fully preserving discrimination ability.

\section{CONCLUSIONS}

In this note, I have illustrated the benefits of score decompositions for a balanced assessment of calibration and predictive performance of probabilistic multi-class classifiers. The results show a substantial decline in predictive performance with recalibration via histogram binning, while isotonic regression produced much better overall results. The smoothed version of isotonic regression with normalization (N-SIR) is found to be particularly effective at improving overall performance, preserving discrimination ability of the base classifiers perfectly while improving calibration in the case study.

In future research, I plan to study the statistical properties of the proposed isotonic estimators including possible biases \citep[cf.][]{FF12,RCSM22}, and extend the study to additional classification tasks and recalibration methods.

\clearpage

\subsubsection*{Acknowledgements}


I would like to thank Tilmann Gneiting and Alexander Jordan, as well as an anonymous reviewer for insightful comments and helpful discussion. The author gratefully acknowledges support from the Deutsche Forschungsgemeinschaft (DFG, German Research Foundation) via project number 502572912.

\subsubsection*{Code and Data Availability}
Code for replicating the case study is available at \url{https://github.com/resinj/calibration_AISTATS26}. Calibration and test data of raw model predictions are available from the supplementary material for \citet{GR22} at \url{https://openreview.net/forum?id=WqoBaaPHS-}.


\bibliography{literature}

\begin{thebibliography}{}

\bibitem[Arnold et~al., 2024]{AWGZ24}
Arnold, S., Walz, E.-M., Ziegel, J.~F., and Gneiting, T. (2024).
\newblock Decompositions of the mean continuous ranked probability score.
\newblock {\em Electronic Journal of Statistics}, 18:4992--5044.

\bibitem[Barlow et~al., 1972]{Barlow1972}
Barlow, R.~E., Bartholomew, D.~J., Bremner, J.~M., and Brunk, H.~D. (1972).
\newblock {\em Statistical Inference Under Order Restrictions: The Theory and
  Application of Isotonic Regression}.
\newblock Wiley, New York.

\bibitem[Berta et~al., 2025]{BHJB25}
Berta, E., Holzmüller, D., Jordan, M.~I., and Bach, F. (2025).
\newblock Rethinking early stopping: Refine, then calibrate.
\newblock Preprint, \href{https://arxiv.org/abs/2501.19195}{arXiv:2501.19195}.

\bibitem[B\l{}asiok et~al., 2023]{BGHN23}
B\l{}asiok, J., Gopalan, P., Hu, L., and Nakkiran, P. (2023).
\newblock A unifying theory of distance from calibration.
\newblock In {\em Proceedings of the 55th Annual ACM Symposium on Theory of
  Computing}, pages 1727--1740.

\bibitem[Brier, 1950]{Brier1950}
Brier, G.~W. (1950).
\newblock Verification of forecasts expressed in terms of probability.
\newblock {\em Monthly Weather Review}, 78:1--3.

\bibitem[Br{\"o}cker, 2009]{Bro09}
Br{\"o}cker, J. (2009).
\newblock Reliability, sufficiency, and the decomposition of proper scores.
\newblock {\em Quarterly Journal of the Royal Meteorological Society},
  135:1512--1519.

\bibitem[Chidambaram and Ge, 2025]{CG25}
Chidambaram, M. and Ge, R. (2025).
\newblock Reassessing how to compare and improve the calibration of machine
  learning models.
\newblock In {\em The Thirteenth International Conference on Learning
  Representations}.

\bibitem[Dimitriadis et~al., 2021]{DGJ21}
Dimitriadis, T., Gneiting, T., and Jordan, A.~I. (2021).
\newblock Stable reliability diagrams for probabilistic classifiers.
\newblock {\em Proceedings of the National Academy of Sciences of the United
  States of America}, 118:e2016191118.

\bibitem[Dimitriadis et~al., 2024]{DGJV24}
Dimitriadis, T., Gneiting, T., Jordan, A.~I., and Vogel, P. (2024).
\newblock Evaluating probabilistic classifiers: The triptych.
\newblock {\em International Journal of Forecasting}, 40:1101--1122.

\bibitem[Ferro and Fricker, 2012]{FF12}
Ferro, C.~A. and Fricker, T.~E. (2012).
\newblock A bias-corrected decomposition of the {B}rier score.
\newblock {\em Quarterly Journal of the Royal Meteorological Society},
  138:1954--1960.

\bibitem[Gneiting and Raftery, 2007]{GR07}
Gneiting, T. and Raftery, A.~E. (2007).
\newblock Strictly proper scoring rules, prediction, and estimation.
\newblock {\em Journal of the American Statistical Association}, 102:359--378.

\bibitem[Gneiting and Resin, 2023]{EJS23}
Gneiting, T. and Resin, J. (2023).
\newblock Regression diagnostics meets forecast evaluation: {C}onditional
  calibration, reliability diagrams, and coefficient of determination.
\newblock {\em Electronic Journal of Statistics}, 17:3226--3286.

\bibitem[Gneiting et~al., 2023]{ARSIA23}
Gneiting, T., Wolffram, D., Resin, J., Kraus, K., Bracher, J., Dimitriadis, T.,
  Hagenmeyer, V., Jordan, A.~I., Lerch, S., Phipps, K., and Schienle, M.
  (2023).
\newblock Model diagnostics and forecast evaluation for quantiles.
\newblock {\em Annual Review of Statistics and Its Application}, 10:597--621.

\bibitem[Gruber and Buettner, 2022]{GB22}
Gruber, S.~G. and Buettner, F. (2022).
\newblock Better uncertainty calibration via proper scores for classification
  and beyond.
\newblock In {\em Advances in Neural Information Processing Systems}, 35.

\bibitem[Guo et~al., 2017]{GPW17}
Guo, C., Pleiss, G., Sun, Y., and Weinberger, K.~Q. (2017).
\newblock On calibration of modern neural networks.
\newblock In {\em Proceedings of the 34th International Conference on Machine
  Learning}, pages 1321--1330.

\bibitem[Gupta and Ramdas, 2022]{GR22}
Gupta, C. and Ramdas, A. (2022).
\newblock Top-label calibration and multiclass-to-binary reductions.
\newblock In {\em International Conference on Learning Representations}.

\bibitem[Huang et~al., 2017]{Huang2017}
Huang, G., Liu, Z., van~der Maaten, L., and Weinberger, K.~Q. (2017).
\newblock Densely connected convolutional networks.
\newblock In {\em Proceedings of the IEEE Conference on Computer Vision and
  Pattern Recognition}.

\bibitem[Hüllermeier and Waegeman, 2021]{HW21}
Hüllermeier, E. and Waegeman, W. (2021).
\newblock Aleatoric and epistemic uncertainty in machine learning: an
  introduction to concepts and methods.
\newblock {\em Machine Learning}, 110:457--506.

\bibitem[Jiang et~al., 2011]{JOKO11}
Jiang, X., Osl, M., Kim, J., and Ohno-Machado, L. (2011).
\newblock Smooth isotonic regression: a new method to calibrate predictive
  models.
\newblock In {\em AMIA Joint Summits on Translational Science Proceedings},
  pages 16--20.

\bibitem[Krizhevsky, 2009]{CIFAR}
Krizhevsky, A. (2009).
\newblock Learning multiple layers of features from tiny images.
\newblock Technical report, University of Toronto.

\bibitem[Kumar et~al., 2019]{KLM19}
Kumar, A., Liang, P.~S., and Ma, T. (2019).
\newblock Verified uncertainty calibration.
\newblock In {\em Advances in Neural Information Processing Systems}, 32.

\bibitem[Machado et~al., 2024]{MCF+24}
Machado, A.~F., Charpentier, A., Flachaire, E., Gallic, E., and Hu, F. (2024).
\newblock Probabilistic scores of classifiers, calibration is not enough.
\newblock Preprint, \href{https://arxiv.org/abs/2408.03421}{arXiv:2408.03421}.

\bibitem[Mukhoti et~al., 2020]{Mukhoti2020}
Mukhoti, J., Kulharia, V., Sanyal, A., Golodetz, S., Torr, P. H.~S., and
  Dokania, P.~K. (2020).
\newblock Calibrating deep neural networks using focal loss.
\newblock In {\em Advances in Neural Information Processing Systems}, 33.

\bibitem[Murphy, 1973]{Murphy73}
Murphy, A.~H. (1973).
\newblock A new vector partition of the probability score.
\newblock {\em Journal of Applied Meteorology and Climatology}, 12:595--600.

\bibitem[Naeini et~al., 2015]{NCH15}
Naeini, M.~P., Cooper, G., and Hauskrecht, M. (2015).
\newblock Obtaining well calibrated probabilities using {B}ayesian binning.
\newblock In {\em Proceedings of the AAAI Conference on Artificial
  Intelligence}.

\bibitem[Nixon et~al., 2019]{NDZ+19}
Nixon, J., Dusenberry, M.~W., Zhang, L., Jerfel, G., and Tran, D. (2019).
\newblock Measuring calibration in deep learning.
\newblock In {\em Proceedings of the IEEE/CVF Conference on Computer Vision and
  Pattern Recognition Workshops}.

\bibitem[Pohle, 2020]{Pohle2020}
Pohle, M.-O. (2020).
\newblock The {M}urphy decomposition and the calibration-resolution principle:
  {A} new perspective on forecast evaluation.
\newblock Preprint, \href{https://arxiv.org/abs/2005.01835}{arXiv:2005.01835}.

\bibitem[Popordanoska et~al., 2024]{PGS+24}
Popordanoska, T., Gruber, S.~G., Tiulpin, A., Buettner, F., and Blaschko, M.~B.
  (2024).
\newblock Consistent and asymptotically unbiased estimation of proper
  calibration errors.
\newblock In {\em Proceedings of The 27th International Conference on
  Artificial Intelligence and Statistics}, pages 3466--3474.

\bibitem[Roelofs et~al., 2022]{RCSM22}
Roelofs, R., Cain, N., Shlens, J., and Mozer, M.~C. (2022).
\newblock Mitigating bias in calibration error estimation.
\newblock In {\em Proceedings of The 25th International Conference on
  Artificial Intelligence and Statistics}, pages 4036--4054.

\bibitem[Rossellini et~al., 2025]{RSB+25}
Rossellini, R., Soloff, J.~A., Barber, R.~F., Ren, Z., and Willett, R. (2025).
\newblock Can a calibration metric be both testable and actionable?
\newblock In {\em Proceedings of Thirty Eighth Conference on Learning Theory},
  pages 4937--4972.

\bibitem[Sanders, 1963]{Sanders1963}
Sanders, F. (1963).
\newblock On subjective probability forecasting.
\newblock {\em Journal of Applied Meteorology and Climatology}, 2:191--201.

\bibitem[Silva~Filho et~al., 2023]{Silva2023}
Silva~Filho, T., Song, H., Perello-Nieto, M., Santos-Rodriguez, R., Kull, M.,
  and Flach, P. (2023).
\newblock Classifier calibration: {A} survey on how to assess and improve
  predicted class probabilities.
\newblock {\em Machine Learning}, 112:3211--3260.

\bibitem[Vaicenavicius et~al., 2019]{VWA+19}
Vaicenavicius, J., Widmann, D., Andersson, C., Lindsten, F., Roll, J., and
  Sch\"{o}n, T.~B. (2019).
\newblock Evaluating model calibration in classification.
\newblock In {\em Proceedings of the 22nd International Conference on
  Artificial Intelligence and Statistics}, pages 3459--3467.

\bibitem[Zadrozny and Elkan, 2002]{ZE02}
Zadrozny, B. and Elkan, C. (2002).
\newblock Transforming classifier scores into accurate multiclass probability
  estimates.
\newblock In {\em Proceedings of the Eighth ACM SIGKDD International Conference
  on Knowledge Discovery and Data Mining}, pages 694--699.

\bibitem[Zagoruyko and Komodakis, 2016]{Zagoruyko2016}
Zagoruyko, S. and Komodakis, N. (2016).
\newblock Wide residual networks.
\newblock In {\em Proceedings of the British Machine Vision Conference}, pages
  87.1--87.12.

\end{thebibliography}

\clearpage
\appendix
\thispagestyle{empty}

\onecolumn
\aistatstitle{Appendix}


\section{RESULTS WITH NORMALIZED SCORE DECOMPOSITION}
\label{App:SDwN}

\begin{figure}[ht]
	\centering
	\includegraphics[width = 0.49\textwidth]{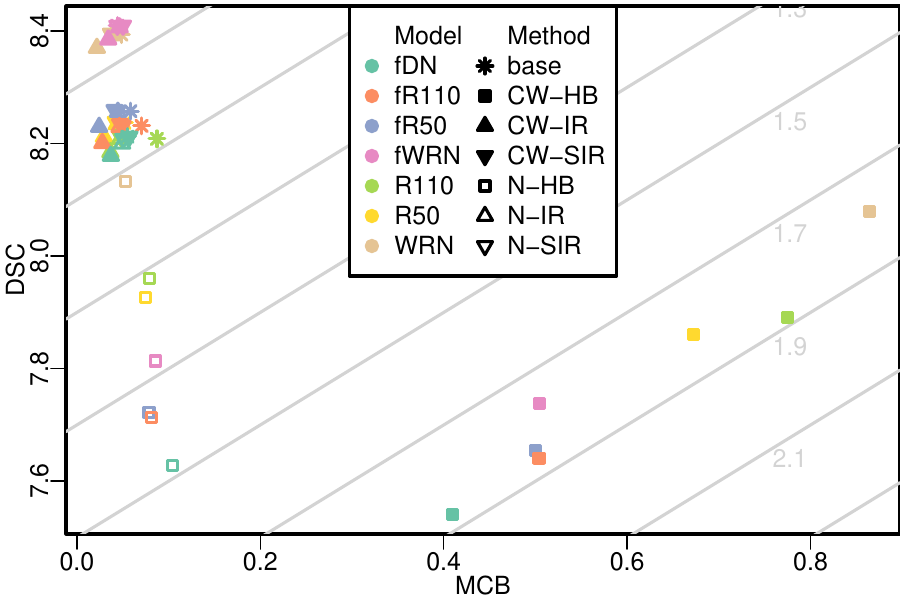}
	\includegraphics[width = 0.49\textwidth]{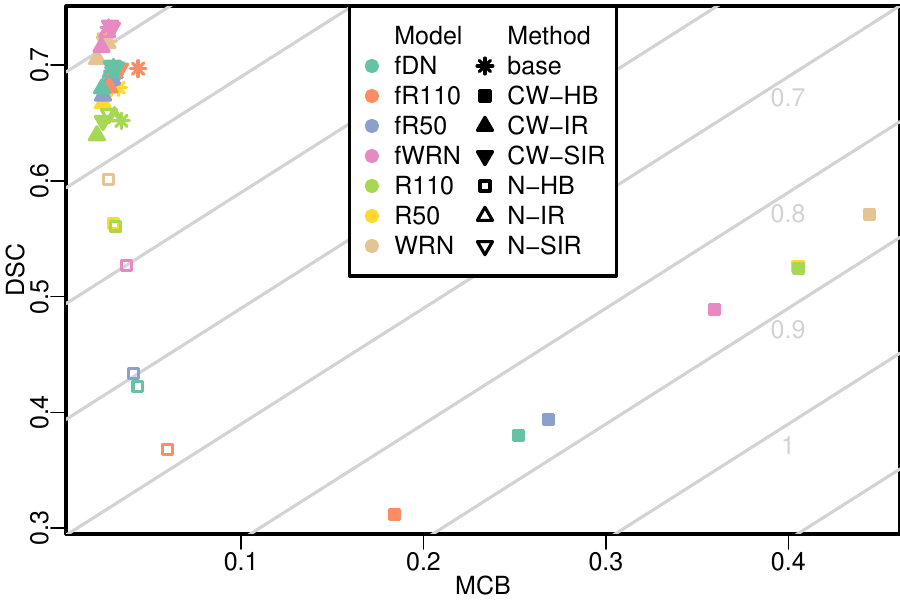}
	\caption{MCB-DSC plots for CIFAR-10 (left) and CIFAR-100 (right) showing \emph{normalized}\/ Brier score decompositions ($\times 100$). The plots show results for seven base models that differ in architecture (\textbf{R}esNet-\textbf{50}, \textbf{R}esNet-\textbf{110}, \textbf{W}ide\textbf{R}es\textbf{N}et, and \textbf{D}ense\textbf{N}et-121) and training loss (\textbf{f}ocal loss or Brier score). The \textcolor{gray}{gray} isolines correspond to MCB-DSC pairs that yield the same overall Brier score.}\label{Fig:MCB-DSC_norm}
\end{figure}

Table \ref{Tab:avgSD_norm} and Figure \ref{Fig:MCB-DSC_norm} present results analogously to Section \ref{Sec:Study} for an alternative \emph{normalized}\/ Brier score decomposition obtained by separately recalibrating each class probability using isotonic regression and then normalizing the recalibrated class probabilities to obtain estimates of the conditional distributions $\widehat{p}_1,\dots,\widehat{p}_n$ used in \eqref{Eq:EmpDecomp}. As alluded to in the main text, conclusions are similar with the exception of the results for CW-HB, where much of the drop in overall score is attributed to miscalibration, because this decomposition quantifies deviations from the stronger notion of calibration in \eqref{Eq:AC} instead of class-wise calibration.

\begin{table}[ht]
	\caption{\emph{Normalized}\/ Brier score decomposition ($\times 100$) and classification accuracy (Acc) averaged across seven base models for both datasets (CIFAR-10 and CIFAR-100) and each recalibration method. The best average values per metric are highlighted in \textbf{bold}. Values in \textcolor{BrickRed}{red} indicate a drastic decline in predictive performance.}\label{Tab:avgSD_norm}
	\centering
	\begin{tabular}{lccccc}
		\toprule
		\multicolumn{6}{c}{CIFAR-10} \\
		\midrule
		Method & $\BS$ & $\MCB$ & $\DSC$ & $\UNC$ & Acc \\ 
		\midrule
		base & 0.781 & 0.059 & 8.279 & 9 & \textbf{0.951} \\ 
		CW-HB & \textcolor{BrickRed}{1.832} & \textcolor{BrickRed}{0.604} & \textcolor{BrickRed}{7.772} & 9 & \textcolor{BrickRed}{0.890} \\ 
		N-HB & \textcolor{BrickRed}{1.237} & 0.080 & \textcolor{BrickRed}{7.842} & 9 & \textcolor{BrickRed}{0.890} \\ 
		CW-IR & 0.779 & \textbf{0.030} & 8.251 & 9 & 0.950 \\ 
		N-IR & 0.770 & 0.045 & 8.275 & 9 & 0.950 \\ 
		CW-SIR & 0.769 & 0.048 & 8.279 & 9 & 0.950 \\ 
		N-SIR & \textbf{0.764} & 0.044 & \textbf{8.280} & 9 & 0.950 \\
		\bottomrule
	\end{tabular}~~
	\begin{tabular}{lccccc}
		\toprule
		\multicolumn{6}{c}{CIFAR-100} \\
		\midrule
		Method & $\BS$ & $\MCB$ & $\DSC$ & $\UNC$ & Acc \\ 
		\midrule
		base & 0.327 & 0.032 & 0.696 & 0.99 & \textbf{0.775} \\ 
		CW-HB & \textcolor{BrickRed}{0.865} & \textcolor{BrickRed}{0.331} & \textcolor{BrickRed}{0.457} & 0.99 & \textcolor{BrickRed}{0.483} \\ 
		N-HB & \textcolor{BrickRed}{0.532} & 0.038 & \textcolor{BrickRed}{0.497} & 0.99 & \textcolor{BrickRed}{0.483} \\ 
		CW-IR & 0.333 & \textbf{0.024} & 0.680 & 0.99 & 0.770 \\ 
		N-IR & 0.324 & 0.029 & 0.694 & 0.99 & 0.770 \\ 
		CW-SIR & 0.323 & 0.029 & 0.696 & 0.99 & 0.773 \\ 
		N-SIR & \textbf{0.320} & 0.028 & \textbf{0.699} & 0.99 & 0.773 \\ 
		\bottomrule
	\end{tabular}
\end{table}

\section{SMOOTHING OF ISOTONIC REGRESSION FITS}
\label{App:SIR}

This section provides a brief description of the employed smoothing. Let $(p_1,y_1),\dots,(p_n,y_n)$ be a sample of binary class probabilities and corresponding binary outcomes encoded as 0 or 1, where $p_i$ is the predicted probability of the outcome $y_i = 1$, and assume w.l.o.g.\ that the sample is ordered by the the predictions, i.e., $p_1 \leq p_2 \leq \dots \leq p_n$. Isotonic regression via the pool-adjacent violators algorithm returns a partition of the sample point indices into consecutive pools $B_1,\dots,B_m \subset \{1,\dots,n\}$ with separating indices $1 = i_1 < i_2 < \dots < i_m < i_{m+1} = n+1$ such that $B_j = \{i_j,i_j + 1, \dots, i_{j+1}-1\}$. For pool $B_j$, let $\bar{y}_j = \frac{1}{\vert B_j \vert}\sum_{i \in B_j} y_i$ be the respective fitted value and $m_j$ be a median of the predictions in the pool, i.e., any value such that $\vert\{p_i \in B_j \mid p_i < m_j\}\vert / \vert B_j \vert \leq \frac12 \leq \vert\{p_i \in B_j \mid p_i \leq m_j\}\vert  / \vert B_j \vert$. Additionally, set $\bar{y}_0 = m_0 = 0$ and $\bar{y}_{m+1} = m_{m+1} = 1$. Then the smoothed isotonic regression fit at $p \in [m_j,m_{j+1}]$ is given by
\[
	f(p) = \frac{\bar{y}_{j+1} - \bar{y}_{j}}{m_{j+1} - m_{j}} (p - m_j) + \bar{y}_j,
\]
which is simply the piecewise linear interpolator of the points $(m_0,\bar{y}_0),\dots,(m_{m+1},\bar{y}_{m+1})$.
In contrast, \citet[Algorithm 1]{JOKO11} ``sample'' the anchor points $m_j$ from the intervals $[\min_{i\in B_j} p_i,\max_{i\in B_j} p_i]$ and apply piecewise cubic Hermite interpolation to obtain a smoothed fit. The resulting function is differentiable, while the simple piecewise linear interpolation only makes the regression function strictly increasing on $[p_1,p_n]$ to avoid mapping distinct predictions to a single recalibrated value.

\end{document}